\documentclass{article}
\usepackage{spconf,amsmath,graphicx}
\usepackage{xcolor}
 \usepackage{amsfonts}
 \usepackage{multirow}
 \usepackage{balance}

\title{BOOSTING DEEP TRANSFER LEARNING FOR COVID-19 CLASSIFICATION}
\name{Fouzia Altaf$^1$, Syed M.S. Islam$^1$, Naeem K. Janjua$^1$ and Naveed Akhtar$^2$ 
\thanks{This work was supported by Australian Government Research Training Program Scholarship.}}
\address{$^1$School of Science, Edith Cowan University, \\$^2$Department of Computer Science, University of Western Australia.}

\begin{document}
%
\maketitle
\begin{abstract}
COVID-19 classification using chest Computed Tomography (CT) has been found pragmatically useful by several studies. Due to the lack of annotated samples,  these studies recommend  transfer learning and explore the choices of pre-trained models and data augmentation. 
However, it is still unknown if there are better strategies than vanilla transfer learning for more accurate COVID-19 classification with limited CT data. This paper provides an affirmative answer, devising a  novel `model' augmentation technique that allows a considerable performance boost to transfer learning for the task. 
Our method systematically reduces the distributional shift between the source and target domains and considers augmenting deep learning with complementary representation learning techniques. We establish the efficacy of our method with  publicly available datasets and models, along with identifying contrasting observations in the previous studies.      


\end{abstract}

\vspace{-3mm}
\section{Introduction}
\label{sec:intro}
\vspace{-2mm}
COVID-19 classification with images is receiving  increasing attention~\cite{yang2020role}, with Computed Tomography (CT) as the leading modality to leverage the super-human predictive abilities of  deep learning~\cite{lecun2015deep} for this critical task~\cite{kundu2020might}, \cite{li2020ct}. CT scans  are widely used for assessing the severity and progression of lung infections~\cite{pham2020comprehensive}. This makes reliable computer aided predictions with CT scans  highly relevant to eventually curb COVID-19. Consequently, there have been multiple studies to explore the practices to maximize deep learning performance for this task. Considering the current lack of clean annotated data, transfer learning with   ImageNet~\cite{imagenet_cvpr09} pre-trained models is the most widely adopted strategy in the current literature. 

Zhao et al.~\cite{zhao2020covid} provided a baseline for COVID-19 classification with public CT-scan images, employing transfer learning. Similarly, \cite{ardakani2020application} uses transfer learning to report results for ten pre-trained models on a dataset of 106 COVID-19 and 86 non-COVID-19 patients. The results are provided using the images pre-processed for regions of interest identification. 
Building on the pre-trained ResNet50~\cite{he2016deep}, Dadario et al.~\cite{dadario2020regarding} proposed COVNet to detect COVID-19 using 4,356 3D CT scans of 3,322 patients. More examples of employing natural image-based pre-trained  deep visual models for COVID-19 detection with CT-scans can also be found, e.g.~\cite{wang2020deep}, \cite{xu2020deep}. 

Except for a very few, e.g.~\cite{zhao2020covid}, the datasets used by the existing works are private. Moreover, the requirement of pre-processing for the region of interest extraction makes their techniques less attractive. Pham~\cite{pham2020comprehensive} provided a comprehensive study of transfer learning for 16 ImageNet  models using a public dataset~\cite{zhao2020covid}. Besides reporting DenseNet201~\cite{huang2017densely} as a promising  architecture for the task, Pham also reported that data augmentation often has a deteriorating effect on vanilla transfer learning for the problem. This finding further caps the training data size for the task, where the correctly annotated data is already limited.

To circumvent the above issue, we investigate if it is possible to augment the classification `model' (instead of training data) to boost COVID-19 classification performance under transfer learning. We provide an affirmative answer to this question with the help of a technique that leverages the fundamentals of machine learning for the performance gain. 
Our method focuses on systematically reducing the distributional shift between the pre-trained model of natural images~\cite{imagenet_cvpr09} and COVID-19 CT-scan images. Moreover, we view deep learning from the lens of representation learning, and augment the overall prediction model with sparse~\cite{tosic2011dictionary} and dense collaborative representation learning~\cite{akhtar2017efficient}. 
We demonstrate that our technique is able to considerably boost the accuracy of COVID-19 classification with limited training data. 

\begin{figure*}
    \centering
    \includegraphics[width = 0.85\textwidth]{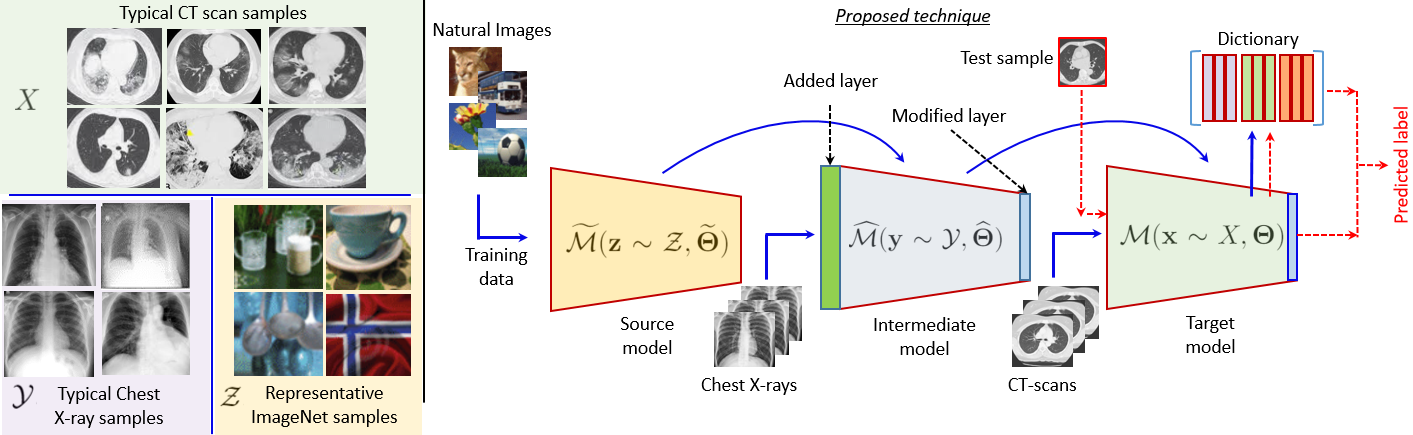}
    \caption{(Left) Typical samples of CT scans, Chest X-ray and ImageNet dataset~\cite{imagenet_cvpr09}. Besides being colored, the patterns in ImageNet samples are very different from CT-scans, identifying a large distributional shift between the domains. The distributional shift is expected to be much smaller between CT scans and X-rays due to the apparent similarities in images, besides both being gray-scale domains. (Right) We propose to first transfer an imageNet model $\widetilde{\mathcal M}(.)$ to an intermediate model $\widehat{\mathcal M}(.)$ of Chest X-rays by adding extra layers that can process gray-scale images. We transfer $\widetilde{\mathcal M}(.)$ to $\widehat{\mathcal M}(.)$ with a relatively large amount of data~\cite{wang2017chestx}. Then, we transfer $\widehat{\mathcal M}(.)$ to the target model ${\mathcal M}(.)$ with the available small COVID-19 CT scan data. We also augment the predictions with sparse~\cite{tosic2011dictionary} and dense collaborative representations~\cite{akhtar2017efficient}. }
    \label{fig:main}
    \vspace{-3mm}
\end{figure*}

\vspace{-3mm}
\section{Motivation}
\vspace{-2mm}
Before introducing the proposed technique, we first highlight the bottleneck of transfer learning for the CT-scan-based COVID-19 classification, which has still kept researchers from achieving the desired level of accuracy with deep learning.
For the discussion, let us denote a deep neural model as a function $\mathcal{M}({\bf x}\sim\mathcal X, \boldsymbol{\Theta})$, where ${\bf x}$ is a sample of the distribution $\mathcal X$ and $\boldsymbol{\Theta}$ is the set of model parameters, a.k.a.~weights. Under the  classification learning objective, the model aims at encoding the distribution $\mathcal X$, which is possible by optimising $\boldsymbol{\Theta}$ over a considerably large set of samples from $\mathcal X$. If the  sample size is small, $\mathcal{M}$ struggles in modeling $\mathcal X$ faithfully. Transfer learning is then employed, which aims at computing the mapping $\boldsymbol\Psi: \widetilde{\mathcal{M}}({\bf z} \sim \mathcal Z, \widetilde{\boldsymbol{\Theta}}) \rightarrow \mathcal{M}({\bf x}\sim X, \boldsymbol{\Theta})$, where $\widetilde{\mathcal M}(.)$ is the pre-trained model learned from a large number of samples of $\mathcal Z$ and $X$ is a small subset of the observed samples of  $\mathcal X$. Given a fixed $X$, the efficacy of the mapping  $\boldsymbol\Psi$ is mainly governed by the distributional shift $|| \mathcal{Z} - \mathcal{X}||$. The smaller is the shift, the more representative is  $\mathcal M(.)$ of the distribution $\mathcal X$, which is desired for better classification of $\mathcal M(.)$ in $\mathcal X$'s domain.   

Unfortunately, the distributional shift between the colored natural images of ImageNet~\cite{imagenet_cvpr09} and the grey-scale images of CT-scans is too large, see Fig.~\ref{fig:main}(left), which compromises the mapping $\boldsymbol\Psi$. Clearly, increasing the size of $X$ could help because the larger distributional shift  entails a larger $||\widetilde{\boldsymbol{\Theta}} - \boldsymbol{\Theta}||$, which can be accounted for with a more comprehensive representation of $\mathcal{X}$  in $X$. However, \cite{pham2020comprehensive} demonstrates that increasing $X$ synthetically does not help for this task. Under our systematic treatment of the problem, we can remark that the data augmentation techniques used in \cite{pham2020comprehensive} are not able to make $X$ more representative of the distribution $\mathcal X$.

Provided that improving $X$ is implausible, we aim at improving the mapping function itself. Namely, we let $\boldsymbol\Psi: \widetilde{\mathcal M}({\bf z} \sim \mathcal Z, \widetilde{\boldsymbol{\Theta}}) \rightarrow \widehat{\mathcal M}({\bf y} \sim \mathcal Y,  \widehat{\boldsymbol{\Theta}}) \rightarrow \mathcal M({\bf x} \sim X, \boldsymbol{\Theta})$, such that $|| \widehat{\boldsymbol{\Theta}} - \boldsymbol{\Theta}|| \ll || \widetilde{\boldsymbol{\Theta}} -  \boldsymbol{\Theta}||$ and we can still compute a reasonable approximation of $\widehat{\mathcal{M}}(.)$ by transferring $\widetilde{\mathcal M}(.)$ to it, because we can  arrange for a larger number of samples of $\mathcal{Y}$. Thus, we reduce the distribution shift between the  source and target models with an intermediate model that has a smaller shift with the target model, whereas it also allows a better transfer of the source model due to the availability of more training data. We give details of the exact procedure in Sec.~\ref{sec:Method}.


Our second major inspiration  comes from looking at deep visual models from the representation learning viewpoint. The model $\mathcal M(.)$ learns a representation of $\mathcal X$ to map its  samples onto a discriminative feature space for classification.  Incidentally, deep learning is not the only representation learning technique available for that purpose. Sparse~\cite{tosic2011dictionary} and dense collaborative representation~\cite{akhtar2017efficient} have also been used effectively for this task. In contrast to the highly non-linear representation learned by deep learning, these methods focus on linear spaces for data modeling. Hence, one can expect them to augment deep learning with their complementary representations. Our results in Sec.~\ref{sec:Exp} verify this. 

\vspace{-3mm}
\section{Method}
\label{sec:Method}
\vspace{-2mm}
We illustrate the proposed method in Fig.~\ref{fig:main}(right) and describe it below following the provided schematics.

\vspace{1mm}
\noindent{\bf Source model $\widetilde{\mathcal{M}}(.)$:} For the  underlying transfer learning task, we follow the common practice of using natural images as the source domain~\cite{pham2020comprehensive}, \cite{ardakani2020application}, \cite{wang2020deep}. 
The models are pre-trained on 1 million labelled images of ImageNet~\cite{imagenet_cvpr09}, mapping  $224\times224\times3$ color images to $1,000$ class labels.  

\vspace{1mm}
\noindent{\bf Intermediate model $\widehat{\mathcal{M}}(.)$:} Considering that our target domain of CT-scans has `large  grey-scale images', we first introduce slight architectural modifications to $\widetilde{\mathcal{M}}(.)$, while preserving its original weights. Concretely, we enforce a larger single channel input of size $448\times448\time1$ to the model by adding an additional convolutional layer such that the output of this layer is a $224\times224\times3$ tensor. For the modification, our strategy is to keep the hyper-parameters of kernel size and strides similar to the first convolutional layer of the original model, and use three filters to output a 3-channel feature map. We use the original activation functions and employ Batch-Normalisation when the original model used it.  

We aim at training the new layer and also fine-tuning the remaining model for an `intermediate' domain to get the intermediate model $\widehat{\mathcal{M}}(.)$. We choose   chest radiography images as our intermediate domain, that provides large-scale annotate data, Chest-Xray14~\cite{wang2017chestx} for thoracic disease classification. Being grey-scale large medical images, this data domain is  closer to the CT-scan images, see Fig.~\ref{fig:main}(left). From \cite{wang2017chestx}, we select a balanced subset of $775$ images per class for $10$ classes, and alter the output layer of $\widehat{\mathcal{M}}(.)$ to predict those classes. We tune the resulting network in a three-step scheme. 

First, we only learn the newly added input layer and the modified output layer for 5 epochs with a learning rate 0.001 using Adam optimizer. This step is intended for a reasonable initialization only.
We further reduce the learning rate 10 times and fined-tuned these layers for 5 more epochs by augmenting the data with a random rotation in [-7,7] degrees, horizontal flip and cropping. For cropping, we select the central $850\times850$  region of $1024\times1024$ images. The network is fed with $448\times 448 \times 1$ input. In the end, we again reduce the learning rate by 10 and allow 5 more epochs to fine-tune the `complete model' with the augmented data.  Note that, data augmentation here is only used as a regularization mechanism for $\widehat{\mathcal{M}}(.)$ to avoid over-fitting to the intermediate domain.    

\vspace{1mm}
\noindent{\bf Target model $\mathcal{M(.)}$:}
To transfer $\widehat{\mathcal{M}}(.)$ to the target domain of CT-scan images, we use $448\times448\times1$ inputs obtained by resizing the CT-scan grey-scale images. Besides the advantage that we  transfer a model of grey-scale medical images to the CT-scan domain, notice that we are also able to use a larger input size (i.e.~$448\times 448$ vs $224\times224$). This is beneficial because larger images contain more information, providing more discriminative patterns. We obtain $\mathcal{M(.)}$ with a further  fine-tuning of $\widehat{\mathcal{M}}(.)$ for 6 epochs with the grey-scale images from the target domain. We use 5e-4 as the learning rate for the whole model, except for the output layer for which the rate is $10\times$ 5e-4 because that layer is added anew to account for the binary classification problem at hand.

\vspace{1mm}
\noindent{\bf Sparse \& dense collaborative  representation:}
Sparse representation~\cite{tosic2011dictionary} encodes a sample, say $\boldsymbol{s} \in \mathbb R^{m}$ as a sparse linear combination of a dictionary $\boldsymbol D \in \mathbb R^{m \times n}$, such that $\boldsymbol D \boldsymbol{\alpha} \approx \boldsymbol s$ and $||\boldsymbol{\alpha}||_0 \leq k$, where $||.||_0$ denotes the pseudo-norm of the vector.  The external constraint $||\boldsymbol{\alpha}||_0 \leq k$ does not allow $\boldsymbol\alpha$ to have more than `$k$' non-zero coefficients. Hence, the representation vector $\boldsymbol\alpha$ is sparse. Removing the sparsity constraint, renders  $\boldsymbol\alpha$ dense. In order to make these representations collaborative, we must construct $\boldsymbol D$ such that its columns (i.e.~the basis vectors) form discriminative subspaces for each class label involved in the problem.   

\begin{table*}[h!]
  \begin{center}
 \caption{Classification results results on SC2C~\cite{soares2020sars} and CCD~\cite{zhao2020covid} datasets. `Transfer learning' denotes conventional transfer learning. `Boosted' denotes our boosted transfer learning method, `Boosted + Data Aug.' also augments data during fine tuning. `Combined' indicates Boosted + Data Aug. model combined with dictionary. Bottom four rows report the results for CCD dataset for `Combined'. Percentage accuracy (Acc.), sensitivity (Sens.), specificity (Spec.) are reported with F1-scores.  }
    \label{tab:table1}
    \begin{tabular}{|l|c|c|c|c|c|}\hline
     \multicolumn{1}{|c|}{\textbf{Models (Dataset)}} & \multicolumn{1}{|c|}{\textbf{Method}} & \textbf{Acc. (\%)}&
     \textbf{Sens. (\%)}&
      \textbf{Spec. (\%)}&
      \textbf{F1 score}\\
      \hline
 \multirow{4}{*}{InceptionV3 (SC2C)} & Transfer learning & 67.60$\pm$0.93 &97.69$\pm$1.02 &38.00$\pm$1.20&0.74
$\pm$0.01\\
& Boosted & 76.34
$\pm$1.71
&97.42$\pm$
0.23
&55.60$\pm$
3.2
&0.80$\pm$
0.01
\\& Boosted + Data Aug. & 78.36$\pm$
1.91
&96.88$\pm$
0.62
&60.13$\pm$
4.23
&0.81$\pm$
0.01
\\
&Combined&80.04$\pm$
0.92
&96.34$\pm$
0.40
&64.00$\pm$
2.22
&0.82$\pm$
0.01\\
\hline
 \multirow{4}{*}{ResNet50 (SC2C)} & Transfer learning & 74.46$\pm$ 1.43 &99.45 $\pm$ 0.62&49.86 $\pm$
3.33
&0.79
$\pm$
0.01
\\
& Boosted & 75.00$\pm$
1.32
&98.37$\pm$
0.40
&52.00$\pm$
3.01
&0.79$\pm$
0.01
\\& Boosted + Data Aug. & 78.89$\pm$
0.76
&98.10$\pm$
1.17
&60.00$\pm$
2.22
&0.82$\pm$
0.01
\\ & Combined & 80.24$\pm$
0.40
&97.83$\pm$
1.30
&62.93$\pm$
2.05
&0.83$\pm$
0.01
\\ \hline

      \multirow{4}{*}{DenseNet201 (SC2C)} & Transfer learning & 74.79$\pm$
2.47
&98.91$\pm$
1.24
&51.06$\pm$
5.99
&0.79$\pm$
0.01
\\
& Boosted & 80.10$\pm$
2.48
&98.78$\pm$
1.22
&61.73$\pm$
3.78
&0.83$\pm$
0.02
\\& Boosted + Data Aug. & 81.31$\pm$
1.34
&99.05$\pm$
1.02
&63.86$\pm$
1.89
&0.84$\pm$
0.01
\\
      & Combined & 82.25$\pm$
2.28
&97.56$\pm$
1.46
&67.20$\pm$
5.89
&0.84$\pm$
0.01
\\
 \hline
 \multirow{4}{*}{VGG16 (SC2C)} & Transfer learning & 79.16$\pm$
2.66
&85.63$\pm$
13.13
&72.80$\pm$
14.54
&0.80$\pm$
0.03\\
& Boosted & 80.24$\pm$
2.82
&89.56$\pm$
13.24
&71.06$\pm$
9.20
&0.81$\pm$
0.04\\
& Boosted + Data Aug. & 83.40$\pm$
0.61
&98.37$\pm$
1.86
&68.66$\pm$
2.34
&0.85$\pm$
0.01\\
& Combined & 84.27$\pm$
0.20
&99.05$\pm$
1.02
&69.73$\pm$
1.22
&0.86$\pm$
0.01
\\\hline \hline
InceptionV3 (CCD) &Combined&78.52$\pm$
1.22&88.60$\pm$
5.37&67.14$\pm$
4.51&0.81$\pm$
0.01\\
\hline
ResNet50 (CCD)&Combined&76.34$\pm$
1.76&81.01$\pm$
5.75
&71.07$\pm$
5.87&0.78$\pm$
0.02\\
\hline
DenseNet201 (CCD) &Combined&76.51$\pm$2.32
&79.74$\pm$
1.03&72.85$\pm$
4.80&0.78$\pm$
0.01\\
\hline
VGG16 (CCD) &Combined&77.26$\pm$
1.26&87.45$\pm$
12.05&66.88$\pm$
11.67&0.79$\pm$
0.02\\
\hline

    \end{tabular}
  \end{center}
  \vspace{-4mm}
\end{table*}

We treat the activation vector before the logits of our final  model as a basis vector for $\boldsymbol D$. Extracting these vectors for the training samples and arranging them in a class-wise manner in a matrix form constructs $\boldsymbol D$ in our approach. Using that, we compute the sparse representation vector of $\boldsymbol s$ using the well-established Orthogonal Matching Pursuit (OMP) technique~\cite{pati1993orthogonal}. Here, $\boldsymbol s$ is the activation vector of $\mathcal M(.)$ for a test sample. For the dense representation vectors, we let $\boldsymbol{\alpha} = (\boldsymbol{D}^{T} \boldsymbol{D} + \lambda \boldsymbol{I})^{-1} \boldsymbol{D}^{T} \boldsymbol{s}$, where $\boldsymbol{I}$ is an identity matrix and $\lambda$ is a scalar. From the linear algebra viewpoint, the computed $\boldsymbol{\alpha}$ gives us a regularized least square projection of $\boldsymbol{s}$ onto the discriminative subspace formed by $\boldsymbol D$. We fuse the two representation vectors by simply normalizing and adding.

\vspace{1mm}
\noindent{\bf Label prediction:}
The computation of sparse and dense representation is done only at the prediction stage. The fused representation vector for a test sample is further combined with the prediction of the target model $\mathcal M(.)$, for which a simple  strategy is adopted. That is, we first add all the coefficients of the fused representation vector  that belong to the same class. It is possible to identify those because our dictionary is an arranged matrix. Then, we add the resulting vector to the softmax activations of $\mathcal M(.)$. 
The intuition is simple. That is, a representation vector for a sample of a given class normally likes to use the dictionary columns belonging to that class more actively. Thus, the corresponding coefficients of the vector gets higher values, which we can use to amplify the softmax scores of $\mathcal M(.)$. In the end, we choose the maximum augmented softmax score to decide the prediction label.    

\vspace{-3mm}
\section{Evaluation}
\label{sec:Exp}
\vspace{-2mm}
We evaluate our technique on two public datasets for COVID-19 classification using CT-scans. 
The first dataset is, SARS-COV-2-CT (SC2C) database~\cite{soares2020sars}. It  contains a total of $2,482$ CT images, which  includes $1,252$ images of COVID-19 positive cases of $60$ patients and $1,230$ images of $60$ COVID-19 negative patients. The data has been collected from different hospitals in Sao Paulo, Brazil. The second dataset is COVID-CT-Dataset (CCD)~\cite{zhao2020covid}. It consists of 349 CT images of  COVID-19 infected patients and 397 CT images of non-infected patients. 
The image sizes in both datasets vary significantly. However, most of those images are much larger than the $224\times224$ grid size. 

We transfer the popular ImageNet models of Inception-v3, ResNet50, DenseNet201 and VGG16 to our target domain using the training details discussed in the previous section. Table~\ref{tab:table1} summarizes the results of our experiments on the two  datasets. We include the results of vanilla `Transfer learning' as the baseline, which is claimed highly accurate in \cite{pham2020comprehensive}. Results for the `Boosted' transfer learning are achieved by transferring our chest X-ray model, which was altered for the larger grey-scale inputs. We use 5 training epochs with 5e-4 learning rate for this transfer. We can see a consistent large performance gain with this improvement over vanilla transfer learning. For the `Boosted + Data Aug.', we also include data augmentation with random scaling in the range [0.9, 1.1], random translation in the range [-5, 5] and reflection. It is worth noticing that data augmentation generally results in a slight performance gain, which is expected. However, this is different from the findings of \cite{pham2020comprehensive}. We discuss this further in Sec.~\ref{sec:disc}.    

Lastly, the `Combined' results indicate that the proposed sparse and dense collaborative representation is also used to improve the `Boosted+Data Aug.' results. Again, generally, an increasing trend in the performance is observed. We use $50$ as the sparsity threshold for the OMP algorithm~\cite{pati1993orthogonal}, and  $\lambda = 2$ to compute the dense representation vector. These values are selected empirically by cross-validation. Due to space restrictions, we only provide `Combined' results for the CDC dataset, reporting similar trends for the remaining methods. Contrary to \cite{pham2020comprehensive}, our results do not particularly favor DensNet201. Instead, shallower networks seem to have a slight advantage. We report results as averages of five draws from the dataset where random chunks of consecutive images were selected as the test data, which formed 10\% of the overall datasets. We note that this strategy and data division is different from \cite{pham2020comprehensive}. For the CDC dataset, we have consistently observed more than 5\% increase over the vanilla transfer learning with our method across all models.

\vspace{-4mm}
\section{Discussion \& Conclusion}
\label{sec:disc}
\vspace{-2mm}
We introduced a novel method to make transfer learning with deep  models of natural images  much more effective for CT-scan-based COVID-19 classification. However, despite a large accuracy gain across all models, the achieved results on public datasets can still not be categorized `acceptable' for automated prediction of this infection. Our results indicate that larger annotated datasets are still required to achieve that target. Otherwise, human experts should not fully rely on the automated results.   Interestingly, our findings do not align well with the existing claims of very high predictive performance of transfer learning on the same datasets, e.g.~Pham's claim~\cite{pham2020comprehensive} of 96\% accuracy with vanilla transfer learning of ImageNet models on \cite{zhao2020covid}. We conjecture that such studies are over-estimating the performance of transfer learning for this task. The apparent high accuracies seem  to be not due to accurate modeling of COVID-19 features, instead they result from encoding data idiosyncrasies to cause a form of over-fitting to the dataset. This argument is supported by two counter-intuitive observations about such studies. (a) Data-augmentation results in significant performance degradation instead of better generalisation. (b) Deeper models perform better than shallower ones despite the small training data size. 
In our separate experiments, we also observed a large  performance degradation in transfer learning results of \cite{pham2020comprehensive}, by slightly changing the training/testing data selection strategy. We refrain from draw conclusive statements about this observation here, and stress on more careful evaluation of transfer learning for this task by the research community. 

As compared to \cite{pham2020comprehensive}, our analysis does not suffer from counter-intuitive observations. However, it also does not support the notion that highly effective transfer learning from the natural image models is possible with limited number of CT-scans. A further investigation for unbiased and fair evaluation of transfer learning for this task is implicated by our study,  which is planned for the future. However, our method does ascertain the possibility of a considerable performance boost for transfer learning for this task.



\balance
\bibliographystyle{IEEEbib}
\bibliography{refs}

\end{document}